\documentclass[fleqn,10pt]{wlscirep}
\usepackage[utf8]{inputenc}
\usepackage[T1]{fontenc}
\usepackage{tabularx}
\usepackage{multirow}
\usepackage{hyperref}
\usepackage{amsmath}
\usepackage[edges]{forest}
\usepackage{adjustbox} 
\usepackage{float}

\newcommand{\modified}[1]{\textcolor{black}{#1}}

\title{GAITEX: Human motion dataset of impaired gait and rehabilitation exercises using inertial and optical sensors}

\author[1,+]{Andreas Spilz}
\author[1,+]{Heiko Oppel}
\author[2]{Jochen Werner}
\author[3]{Kathrin Stucke-Straub}
\author[2]{Felix Capanni}
\author[1]{Michael Munz}
\affil[1]{AI for Sensor Data Analytics Research Group, Ulm University of Applied Sciences, Ulm, 89081, Germany}
\affil[2]{Biomechatronic Research Group, Ulm University of Applied Sciences, Ulm, 89081, Germany}
\affil[3]{Institute of Computer Science, Ulm University of Applied Sciences, Ulm, 89081, Germany}

\affil[*]{Michael.Munz@thu.de}

\affil[+]{these authors contributed equally to this work}

\begin{abstract}
Wearable inertial measurement units (IMUs) provide a cost-effective approach to assessing human movement in clinical and everyday environments. However, developing the associated classification models for robust assessment of physiotherapeutic exercise and gait analysis requires large, diverse datasets that are costly and time-consuming to collect. We present a multimodal dataset of physiotherapeutic and gait-related exercises, including correct and clinically relevant variants, recorded from 19 healthy subjects using synchronized IMUs and optical marker-based motion capture (MoCap). It contains data from nine IMUs and 68 markers tracking full-body kinematics. Four markers per IMU allow direct comparison between IMU- and MoCap-derived orientations. We additionally provide processed IMU orientations aligned to common segment coordinate systems, subject-specific OpenSim models, inverse kinematics outputs, and visualization tools for IMU-derived orientations. The dataset is fully annotated with movement quality ratings and timestamped segmentations. It supports various machine learning tasks such as exercise evaluation, gait classification, temporal segmentation, and biomechanical parameter estimation. Code for postprocessing, alignment, inverse kinematics, and technical validation is provided to promote reproducibility.

\end{abstract}
\begin{document}

\flushbottom
\maketitle
%
%
\thispagestyle{empty}

\section*{Background \& Summary}
Human motion datasets play a crucial role in the development and validation of computational models for movement analysis, rehabilitation support systems, and activity recognition.
While a number of existing datasets capture general locomotor activity\cite{Luo2020}, \cite{Angelidou2025} or gait under specific conditions\cite{Bacek2024}, there remains a need for datasets addressing physiotherapeutic exercises and gait disorders.
In particular, differentiating between subtle variations in the execution of physiotherapeutic exercises or detecting compensatory mechanisms in impaired gait are key challenges in the development of robust and responsive systems for rehabilitation monitoring.

To address this gap, we introduce a dataset comprising two complementary recording scenarios. 
The first scenario focuses on physiotherapeutic exercises commonly used in the treatment of foot drop, a condition characterized by difficulty in lifting the front part of the foot. 
Healthy participants performed two commonly used rehabilitation exercises under supervision: (1) resisted dorsiflexion and (2) resisted gait simulation. 
Both exercises were executed in four distinct ways; one correct and three intentionally incorrect variations, which were executed under the guidance of an instructor. 
This structure provides a controlled framework for the development and evaluation of classification models that can distinguish between correct and different types of incorrect exercise executions, an essential step in building automated feedback systems. 
Notably, the dataset does not include individuals from clinical populations, as its primary aim is to support methodological development rather than clinical assessment.
The second scenario comprises treadmill walking data designed to investigate gait anomalies.
Participants first walked freely on a treadmill at a self-selected pace that increased incrementally by 0.5 km/h per minute, over a total of three minutes. 
Afterwards, they were then fitted with a knee orthosis that restricted knee flexion to 0° and repeated the walking trial. 
The comparison of natural and constrained gait enables the study of altered biomechanical patterns, which may inform models for anomaly detection or aid design in wearable assistive technologies.

A key feature of this dataset is the synchronized recording of nine body worn inertial measurement units (IMUs) (Xsens Awinda, Movella Holdings Inc., Henderson, Nevada, United States of America) alongside a state of the art optical Motion Capturing (MoCap) system (Qualisys, Göteborg, Sweden). The MoCap system was used to track the participants' full-body kinematics as well as the movement of the each IMU housing. 
This setup enables frame-synchronized, segment-level validation of IMU-derived motion data against marker-based reference trajectories. 
Several research groups have previously employed similar methodologies, but typically under more constrained conditions, with reduced sensor configurations or with a focus on only gait related exercises.
For instance, Santos et al.\cite{Santos2022} recorded gait information using a MoCap system and only two IMUs, both attached to the same segment, thereby limiting the spatial information to effectively a single measurement point.
The study focuses exclusively on healthy gait and therefore offer valuable insights into locomotor behavior. Our dataset extends these efforts by incorporating physiotherapeutic exercises and gait patterns affected by functional impairments.

In addition to the raw recordings, we provide a post-processed version of the IMU data, in which sensor orientations are expressed relative to subject-specific anatomical segment frames. Using OpenSim \cite{seth_opensim_2018}, we further derived joint kinematics via an inverse kinematics workflow based exclusively on these IMU orientations.
This workflow not only yields biomechanically interpretable joint trajectories but also facilitates visual quality control of the reconstructed sensor orientations, enabling researchers to assess the plausibility and internal consistency of the IMU-based data.
In contrast to Scherpereel et al.\cite{Scherpereel2023}, who applied a similar inverse kinematics procedure based on optical marker trajectories, our approach relies solely on wearable sensor data and therefore provides a valuable resource for researchers developing and evaluating IMU-only methods.

We anticipate that this dataset will be of value to researchers working in the areas of rehabilitation technology, biomechanics and sensor-based movement analysis. 
It offers well-labeled, multimodal recordings collected under controlled conditions and is suited for algorithm development, benchmarking and exploration of motor variation in both therapeutic and naturalistic contexts. 

The main contributions of this work include: 
\begin{itemize}
    \item The release of a multimodal dataset \cite{munzDatasetGAITEXHuman2025} comprising raw orientation data from nine body-worn IMUs; time-synchronized MoCap data that track both the participant's body segments and the IMU sensor housings and manually curated temporal annotations delineating individual repetitions of each task from the continuous sensor streams.
    \item A processed version of the IMU data, in which sensor orientations have been transformed into a segment-aligned reference frame using subject-specific calibration, thereby eliminating variability introduced by sensor placement differences across participants.
    \item The provision of personalized, anatomically scaled musculoskeletal models and corresponding inverse kinematics results derived from the IMU signals, enabling joint angle estimation and visualization of IMU orientations.
    \item An open-source processing pipeline that allows users to reproduce and adapt all postprocessing steps, including model scaling and the application of inverse kinematics to raw sensor data.
\end{itemize}

\section*{Methods}

\subsection*{Participants and ethical requirements}
In total, 19 adult volunteers (15 male, 4 female) participated in the present study. All participants were free from any known allergic reactions to Kinesio tapes or adhesives, did not present with acute general illnesses or orthopedic conditions incompatible with participation, and were not pregnant at the time of data acquisition.
The data collection campaign was conducted between September 1st and December 15th, 2024. 
Participants’ ages ranged from 22 to 55 years. For men, the mean age was 39.07 ± 10.44 years (range: 22–55 years), and for women, 34.75 ± 12.97 years (range: 25–53 years).
Participant height ranged from 158 cm to 192 cm. Men had a mean height of 180.53 ± 6.33 cm (range: 171–192 cm), while women had a mean height of 163.75 ± 4.92 cm (range: 158–170 cm).
Body weight varied between 52 kg and 102 kg. The mean body weight was 78.40 ± 11.33 kg for men (range: 62–102 kg) and 61.25 ± 11.64 kg for women (range: 52–78 kg).
The Body Mass Index (BMI) ranged from 19.57 to 29.39 kg/m². For men, the mean BMI was 23.98 ± 2.58 kg/m² (range: 20.91–29.39 kg/m²), and for women, 22.76 ± 3.42 kg/m² (range: 19.57–26.99 kg/m²).

All participants were fully informed about the experimental procedures, potential risks, and the intended use of the collected data. 
Written informed consent was obtained from all individuals prior to participation.
The experimental protocol was reviewed and approved by the Ethics Committee of the Ulm University of Applied Sciences (approval date: April 18, 2024; reference number: 2024-01). The study is registered in the German Clinical Trials Register (DRKS) under the DRKS-ID DRKS00034705.

\subsection*{Acquisition Setup}

The measurements were conducted in the motion analysis laboratory at the Ulm University of Applied Sciences. 
Participants performed the designated movement tasks on an h/p/cosmos treadmill. 
For walking tasks, the treadmill was activated, whereas for physiotherapeutic movements, it remained stationary. 
A safety bar attached to the treadmill in combination with a safety harness served as fall protection during walking activities. 
Additionally, a custom-built apparatus was mounted at the rear end of the treadmill, allowing the attachment of resistance bands with adjustable tension. This setup enabled participants to be positioned under tension, as required for the execution of the targeted physiotherapeutic exercises. The configuration of the resistance band system was adapted to each exercise type and adjusted individually for each participant.

\begin{figure}[htb]
\centering
\includegraphics[width=\linewidth]{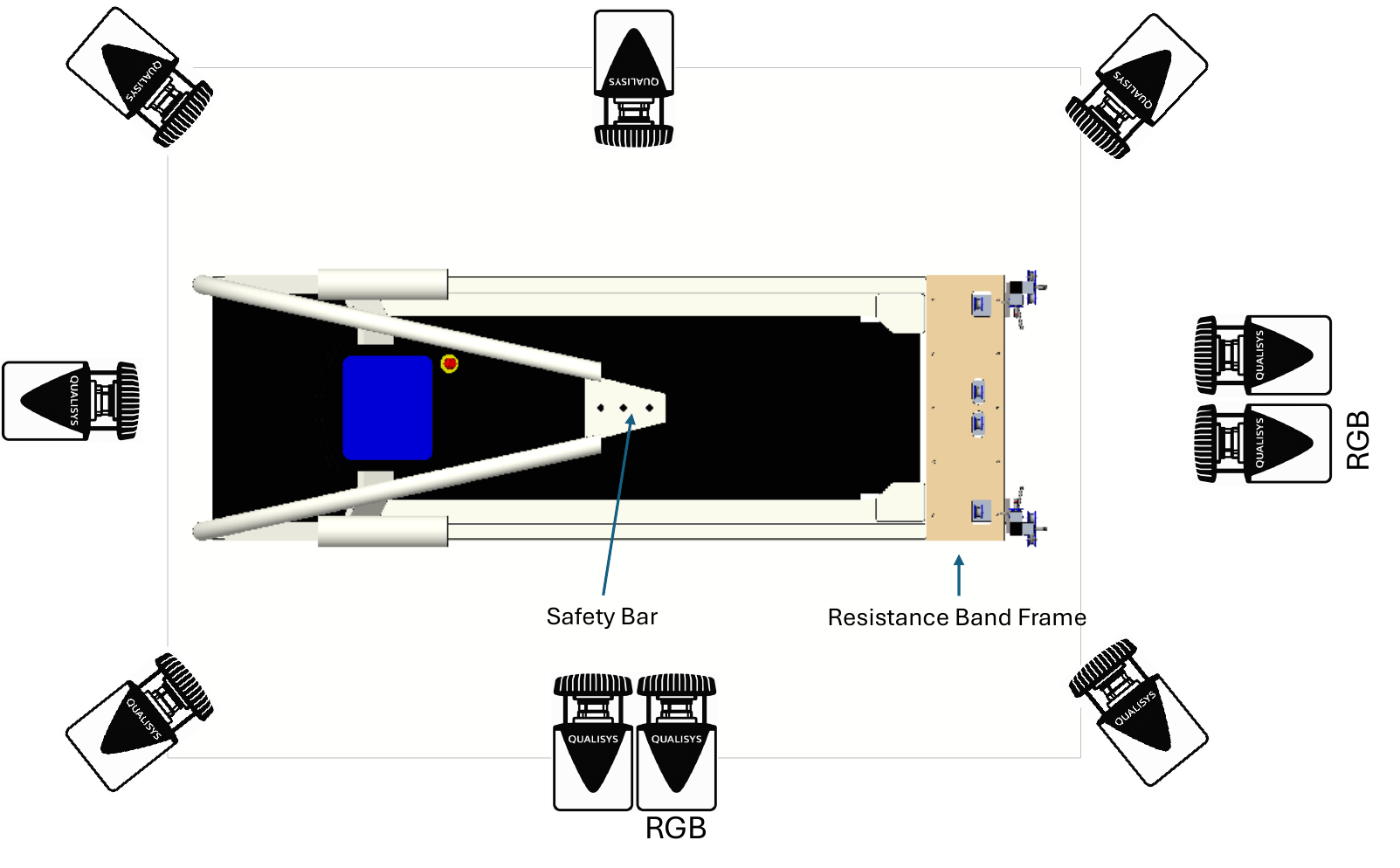}
\caption{Top view of the experimental setup including treadmill, safety equipment, resistance band frame, and camera configuration. Eight Qualisys cameras were used for marker tracking, and two RGB cameras for video recording.}
\label{fig:aqu_setup}
\end{figure}

\begin{figure}[htb]
\centering
\includegraphics[width=\linewidth]{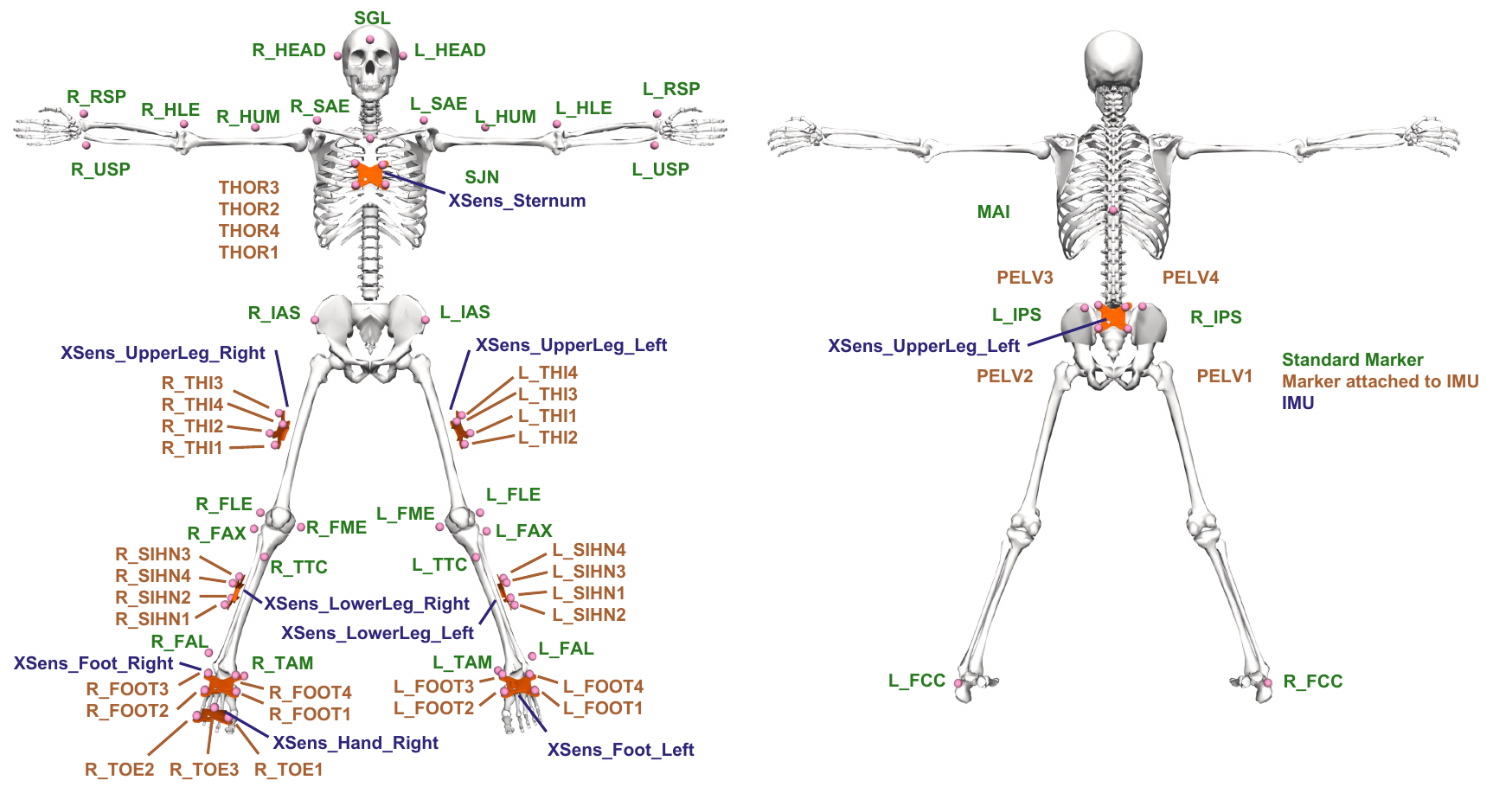}
\caption{Anterior (left) and posterior (right) views of the experimental marker setup. IMUs are shown in blue, Qualisys markers used for segment pose tracking are marked in red, and Qualisys markers mounted on IMUs for optical orientation validation are shown in orange. Marker and sensor labels match the naming conventions used in the dataset. \modified{Note that the IMU placed on the right forefoot is labeled as \texttt{XSens\_Hand\_Right}, since the standard Xsens setup does not include a dedicated IMU for this position. Therefore, the otherwise unused right-hand IMU was repurposed for the right forefoot.} }
\label{fig:marker_imu_placement}
\end{figure}

Around the treadmill, 10 cameras were arranged as part of a Qualisys optical MoCap system. 
Of these, eight cameras were dedicated to marker tracking, while two RGB cameras recorded the performed exercises. 
An overview of the measurement environment is provided in Figure \ref{fig:aqu_setup}. \newline
To achieve comprehensive full-body kinematics, 33 reflective markers were attached to each participant (see Figure \ref{fig:marker_imu_placement}). 
The marker arrangement was based on the established Institute of Orthopaedic Research (IOR) marker set, slightly adapted for our specific application needs. 
These adaptations involved the removal of several markers that conflicted with the placement of IMUs (markers on the toes and markers on the lower back) or essential safety equipment (markers on the upper back the sternum and the fingers), preventing their proper attachment. \newline
The Qualisys MoCap system recorded the spatial trajectories of these markers with the eight mentioned infrared cameras positioned around the participants, operating at a sampling frequency of 100 Hz using the QTM software (v2023.3). 
To ensure high-quality reference data for subsequent analysis, we carefully calibrated the Qualisys motion capture system following the manufacturer’s recommendations. Calibration was only accepted if the standard deviation of wand length measurements remained below 1~mm, indicating stable camera positioning and consistent marker detection across views.\newline
Additionally, two RGB cameras recorded the performed exercises at 25 frames per second. 
The video recordings from these RGB cameras were used later to segment the individual repetitions of exercises and to precisely define the start and end points of individual repetition.

In addition to the MoCap system, the used setup employed nine Xsens MTw Awinda IMUs, which were primarily placed on the lower limbs of each participant, as illustrated in Figure \ref{fig:marker_imu_placement}. 
Each IMU recorded motion data at a sampling rate of 100 Hz using the MT Manager software (v4.6). \newline
Orientation estimation was performed onboard on each sensor using the proprietary Xsens Kalman filter algorithm (XKF3hm), which computes the orientation of the sensor’s local coordinate system (CS) relative to an estimated Earth-fixed reference frame. 
The computed orientation data were transmitted wirelessly to a central recording PC. \newline
To enable validation of the IMU orientation estimates, a custom sensor mount was designed to attach four reflective Qualisys markers directly to each IMU. \modified{The IMU on the right forefoot carried only three markers due to space constraints (see Figure \ref{fig:marker_imu_placement}).}
This configuration allowed the IMU orientation to be independently derived from the optical motion capture system, facilitating a comparative analysis of IMU-based and marker-based orientation estimates.

Temporal synchronization of the measurement systems was ensured through sync units provided by Xsens and Qualisys. 
The Xsens sync unit acted as the master device, supplying a measurement frequency and signals marking both the beginning and end of each measurement.
This setup ensured precise synchronization and simultaneous data acquisition by both measurement systems.

\subsection*{Acquisition protocol}
\label{section:acqProtocol}
Over a period of four months, 19 participants performed two physiotherapeutic and two gait-related movement tasks while equipped with the described sensor setup. \modified{Each participant performed the movement tasks in a predefined, non-randomised order, which corresponds to the sequence listed in Table 1.}

\begin{figure}[htb]
\centering
\includegraphics[width=\linewidth]{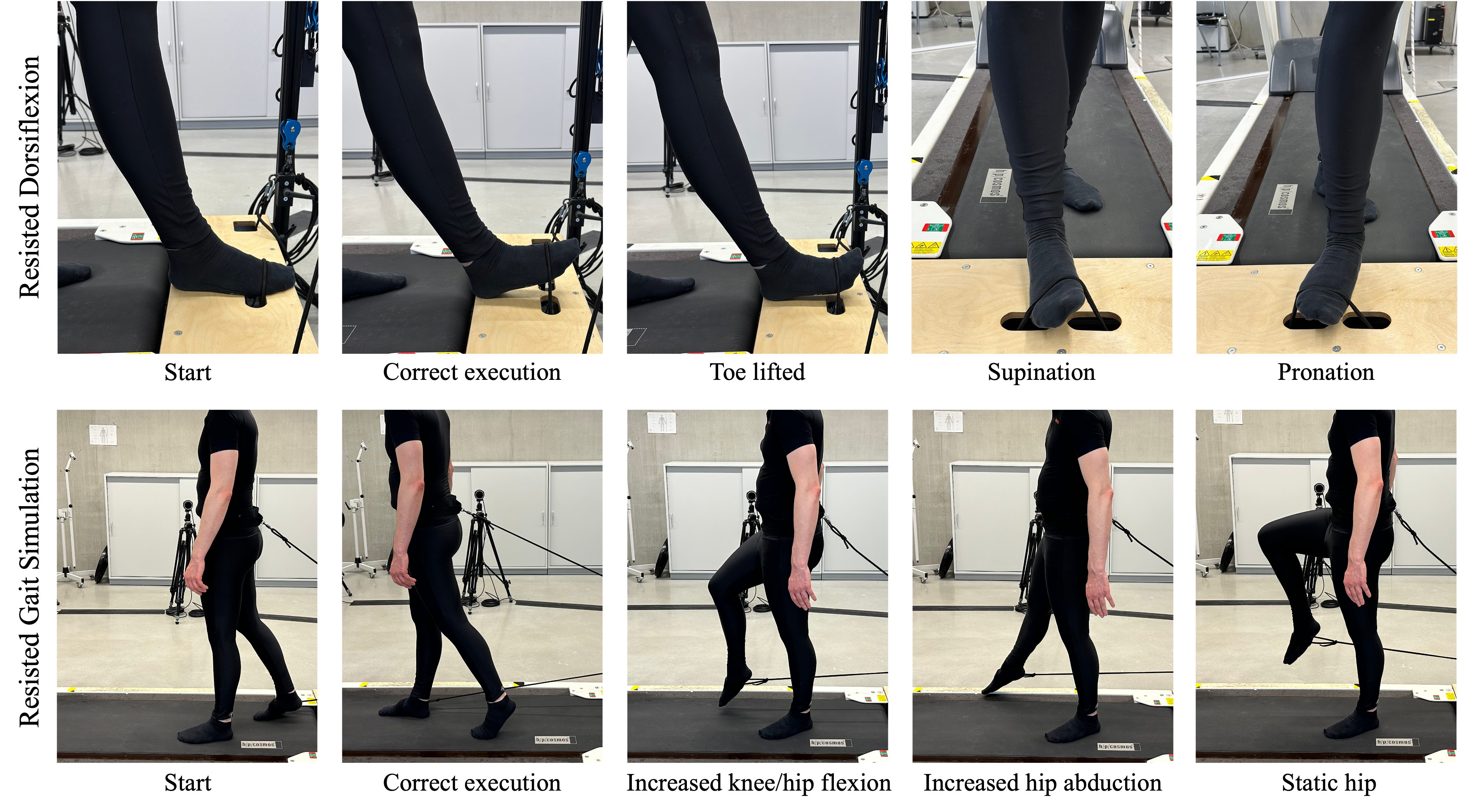}
\caption{Overview of the physiotherapeutic exercises and their execution variants. \modified{Both exercises are performed in a standing position.} The first exercise RD is illustrated in the upper row, and the second exercise RGS is illustrated in the lower row. For each exercise, the first image from the left illustrates the standardized start position. The second image depicts the correct execution according to physiotherapeutic guidelines. The remaining three images on the right display common erroneous movement patterns observed in individuals with foot drop. Detailed descriptions of the respective deviations can be found in Table \ref{tab:exercise_protocol}.}
\label{fig:phy_ex}
\end{figure}

The physiotherapeutic exercises were selected from treatments commonly employed for addressing foot drop. 
The first physiotherapeutic exercise "Resisted Dorsiflexion" (RD) aimed at strengthening the muscles involved, particularly emphasizing the dorsiflexion of the foot. 
An adjustable resistance band was used to tailor the training intensity for each participant (depicted in Figure \ref{fig:phy_ex}). \modified{The tension was adjusted according to the participant’s individual strength and range of motion by the supervisors.} \newline
The second physiotherapeutic exercise "Resisted Gait Simulation" (RGS) aimed to induce training effects within the context of a natural gait cycle. 
The core task involved executing half a gait cycle, with resistance bands providing additional muscular stimulation to improve gait stability and alleviate foot drop (depicted in Figure \ref{fig:phy_ex}).\newline
To gather sufficient data for identifying typical errors in exercise execution, healthy participants without foot drop performed four variations of each exercise during this campaign. 
One variation corresponded to the correct (textbook) execution, while the other three represented common movement errors observed in patients with foot drop. 
The individual variants are shown in Figure \ref{fig:phy_ex} and are further described in Table \ref{tab:exercise_protocol}.
Prior to data collection, participants received structured instruction on all four movement variants. Each variant was first demonstrated by the experimenter, after which participants were given the opportunity to practice the corresponding movement under supervision. 
During this familiarization phase, participants received immediate corrective feedback whenever deviations from the intended execution occurred. 
Recordings were only initiated once participants demonstrated consistent and reliable performance of each variant. 
\modified{To minimise the potential influence of fatigue, participants were instructed to take a brief rest after completing each exercise variant, corresponding to ten repetitions per set.}

\begin{figure}[htb]
\centering
\includegraphics[width=10cm]{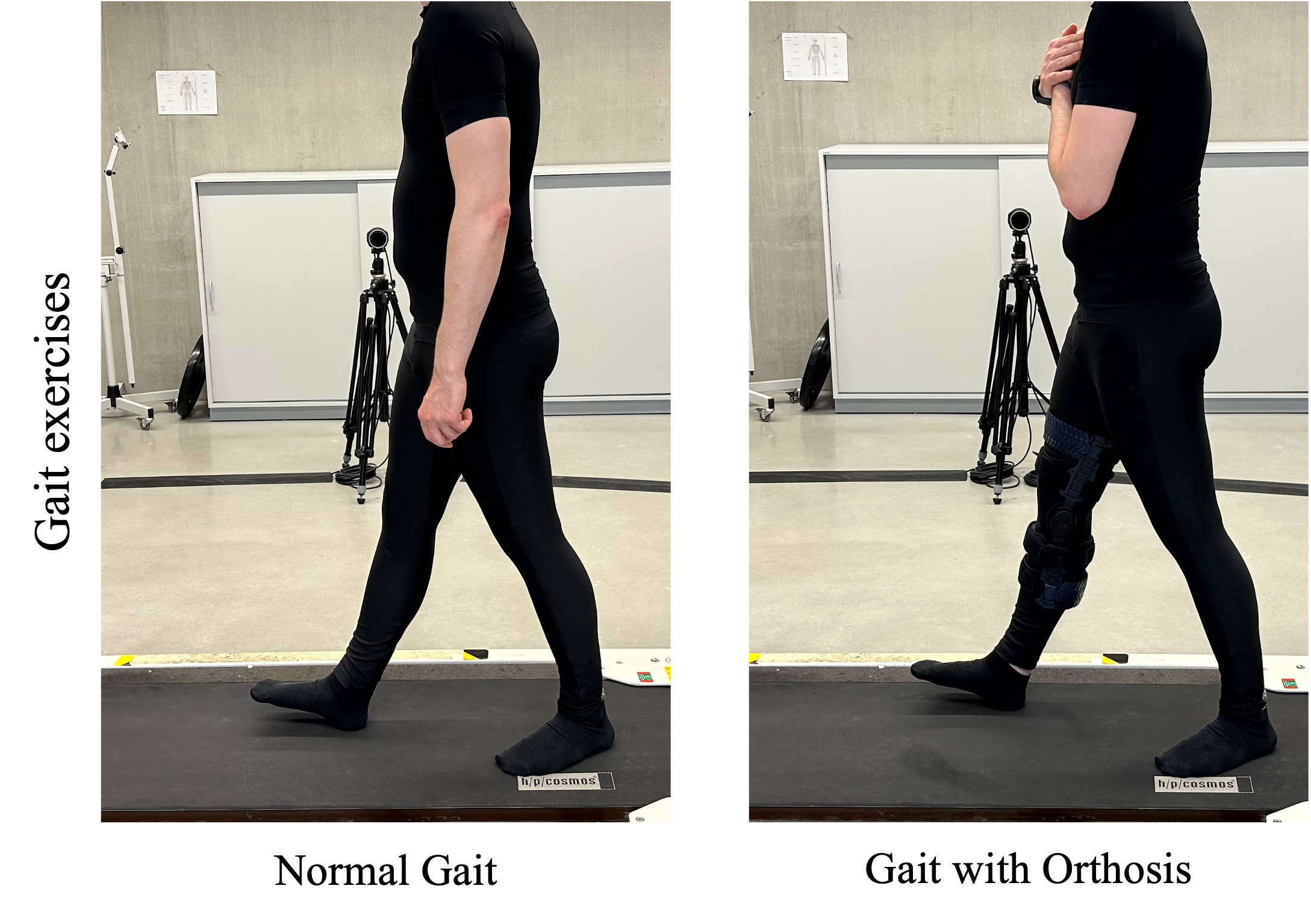}
\caption{Illustration of the two gait-related tasks. The left image shows a participant walking on a treadmill during the NG condition, without any assistive device. The right image depicts the same participant performing the GWO task under otherwise identical conditions, while wearing a rigid knee orthosis on the right leg that restricts knee flexion to 0°.}
\label{fig:gait_ex}
\end{figure}

The first gait-related task "Normal Gait" (NG) involved capturing participants' natural walking patterns on a treadmill at three different speeds. 
Given the absence of physical impairments, a natural gait pattern with healthy variations was expected. 
Each participant performed the walking task at three self-selected treadmill speeds, recorded consecutively for approximately one minute each. Starting from an individually chosen baseline speed, the treadmill velocity was increased twice in small increments, resulting in a total speed difference of approximately 1 km/h between the slowest and fastest condition. \modified{The exact duration and magnitude of each increment varied across participants and can be retrieved from the corresponding \texttt{timestamps\_...} file (see Figure~\ref{fig:folder_tree}).} \newline
The second gait-related task "Gait with Orthosis" (GWO) followed an identical protocol (including the same individual chosen treadmill speeds) but required participants to wear a knee orthosis on the right knee, restricting knee flexion to 0° (see Figure \ref{fig:gait_ex}).

At the beginning of each recording session, participants were instructed to assume a standardized reference posture, a T-pose with legs abducted (see Figure \ref{fig:marker_imu_placement}), to facilitate subsequent model calibration. 
After holding this static pose briefly, they proceeded to perform the respective motor task. 
The sequence of exercises followed during the session is detailed in Table \ref{tab:exercise_protocol}, which also provides descriptions of each task and the corresponding aliases used in the dataset.

\begin{table}[htbp]
\centering
\caption{Overview of exercise types and variants used for data collection. Each exercise type is listed with its respective variant, description, recording duration or repetitions, and the alias used in the dataset.}
\begin{tabularx}{\textwidth}{X X X X X}
\toprule
\textbf{Exercise} & \textbf{Variant} & \textbf{Description} & \textbf{Reps/Duration} & \textbf{Alias} \\
\midrule

\multirow{4}{=}{Resisted Dorsiflexion (RD)} 
& Correct execution & Dorsiflexion of the right foot with neutral toe position; no supination or pronation. & 10 reps & rd\_correct \\
& Toe lifted & Initiation of dorsiflexion via toe lift; otherwise correct execution. & 10 reps & rd\_toes \\
& Supination & Dorsiflexion with foot in supination; toes neutral. & 10 reps & rd\_supination \\
& Pronation & Dorsiflexion with foot in pronation; toes neutral. & 10 reps & rd\_pronation \\

\midrule

\multirow{4}{=}{Resisted Gait Simulation (RGS)} 
& Correct execution & Gait from terminal contact to mid-stance on the right foot and reverse; no abnormalities. & 10 reps & rgs\_correct \\
& Increased knee/hip flexion & Drop foot compensated with exaggerated hip and knee flexion. & 10 reps & rgs\_flexion \\
& Increased hip abduction & Drop foot compensated with reduced knee flexion and increased hip abduction. & 10 reps & rgs\_abduction \\
& Static hip & Hip remains static; drop foot compensated with exaggerated leg lift ("stork gait"). & 10 reps & rgs\_stork \\

\midrule

\multirow{3}{=}{Normal Gait (NG)}
& Baseline speed & Walking on treadmill at self-selected speed $v_0$. & $\sim$1 min & ng\_v0 \\
& Increased speed & Walking at $v_1 > v_0$; participant-defined increment. & $\sim$1 min & ng\_v1 \\
& Top speed & Walking at $v_2 > v_1$; participant-defined increment. & $\sim$1 min & ng\_v2 \\

\midrule

\multirow{3}{=}{Gait with Orthosis (GWO)}
& Baseline speed & Walking at $v_0$ with right knee orthosis limiting flexion to 0°. & $\sim$1 min & gwo\_v0 \\
& Increased speed & Walking at $v_1$ with right knee orthosis limiting flexion to 0°. & $\sim$1 min & gwo\_v1 \\
& Top speed & Walking at $v_2$ with right knee orthosis limiting flexion to 0°. & $\sim$1 min & gwo\_v2 \\

\bottomrule
\end{tabularx}
\label{tab:exercise_protocol}
\end{table}

\subsection*{Postprocessing}
\label{section:post}
In the following, we outline the postprocessing procedures applied to the collected dataset.

Notably, temporal synchronization between the Qualisys MoCap system and the Xsens IMUs is achieved by hardware-synchronization using dedicated synchronization units. This ensures temporal alignment during recording.

From the RGB video recordings, task-specific timestamps were manually extracted. For the physiotherapeutic exercises, we annotated the start and end of each individual repetition. 
In contrast, for the gait trials, we documented the start and end times of each walking speed condition.

\modified{The assignment of exercise repetitions to its corresponding variant (e.g. correct, toe lifted, etc.) was based on the verbal instructions given to the participants during data acquisition and subsequently verified through manual inspection of the video data by a supervisor.
Since the labels were derived directly from the instructed exercise variants rather than from independent manual annotation, the labels should be interpreted as reflecting the intended exercise condition rather than a qualitative assessment of the movement execution.}

The raw data from the Xsens IMUs and the optical motion capture system are provided in unaltered numerical form. However, data is not stored in the native export formats of the respective acquisition software.
Instead, all data were converted into a unified, human-readable CSV format to facilitate usability and downstream processing.
During this conversion, no filtering or modification of the underlying measurements were applied. 
However, naming inconsistencies, such as non-standardized IMU identifiers introduced by the original acquisition software, were harmonized to ensure consistency across all recordings.
Marker trajectories from the Qualisys system were likewise preserved in their raw form. 
The assignment of marker labels was performed manually based on the experimental setup and subsequently validated for anatomical plausibility. 
In cases where a marker was not tracked for a certain period, no interpolation or gap-filling was applied. 
Instead, a placeholder value of 0.0 was inserted for the x, y, and z coordinates.
All time series share a common, synchronized time base.

\subsection*{Inverse Kinematics using IMU Data and Musculoskeletal Models}
\label{section:ik}
To enable intuitive visualization and interpretation of the recorded motion data, we performed musculoskeletal simulations using the open-source software OpenSim (v.4.4.1). 
Based on the raw IMU orientation data, we computed joint angle trajectories that allow reconstruction of the captured movements within a biomechanical model. 
This approach facilitates a human-interpretable representation of the recorded exercises, without the need to share video recordings, which may raise privacy concerns and potentially violate participant data protection rights. \newline
In addition to the raw data, we provide all necessary files required to reproduce these simulations and visualizations within OpenSim, including subject-specific scaled models and processed IMU orientation data. 
All processing steps are transparently documented and fully reproducible using the accompanying GitHub repository, which contains code and configuration files to replicate the simulations from raw inputs.\newline 
In the following, we describe the underlying procedure in detail. 
Specifically, marker trajectories from the MoCap system were used to perform subject-specific scaling of the generic musculoskeletal model. 
However, the actual Inverse Kinematics computations were carried out using the IMU orientation data.
This strategy was chosen to enable visualization of orientation-based IMU measurements, which are otherwise difficult to interpret in isolation. 
Nonetheless, users who prefer a marker-based Inverse Kinematics workflow may do so using the provided marker trajectories.

As a first step, a subject-specific scaling of the musculoskeletal model was performed for each recording using the OpenSim \texttt{ScaleTool}. This process adjusts the geometry of each body segment based on the recorded marker positions, thereby tailoring the model to the individual participant’s anthropometry. 
We used a full-body model originally published by Rajagopal et al. \cite{rajagopal_full-body_2016} as the anatomical basis and adapted it to our experimental requirements. 
These adaptations included the integration of our custom Qualisys marker setup and the removal of joint motion constraints to ensure that the recorded IMU-based movements could be visualized without artificial restrictions.
For the scaling process itself, we used the static calibration data obtained during the T-pose, which each participant performed at the beginning of each measurement.
The calculated static posture was defined as the model’s default pose.

In the subsequent step, joint angle trajectories were estimated using the \texttt{IMUInverseKinematicsTool} provided by OpenSim. 
To enable this computation, it is essential to determine the static rotational offset between each IMU and its corresponding body segment in the musculoskeletal model. 
This transformation ensures that the time-varying orientation estimate calculated by the IMUs can be meaningfully interpreted in the context of the model and used as input for the Inverse Kinematics algorithm. 
The required transformation $^{XKF3hm}_{Segment}T$, which maps the orientation estimated by the XKF3hm algorithm to the coordinate frame of the corresponding OpenSim segment (Segment), can be expressed as follows:
\begin{equation}
^{XKF3hm}_{Segment}T = ^{OpenSim}_{Segment}T\cdot^{OpenSim_y}_{OpenSim}T\cdot^{XKF3hm}_{OpenSim_y}T
\label{eq:transf}
\end{equation}

Each component of the transformation is defined as follows: \newline
$^{XKF3hm}_{OpenSim_y}T$ is a fixed transformation aligning the y-axis of the XKF3hm CS with that of the OpenSim CS, ensuring consistent vertical orientation.\newline
$^{OpenSim_y}_{OpenSim}T$ accounts for rotational differences between the two CS around the vertical axis. 
It is computed by determining the heading / yaw angle (rotation about the y-axis) between the initial orientation of the IMU and the orientation derived from the Qualisys markers mounted directly on the IMU.\newline
$^{OpenSim}_{Segment}T$ defines the transformation between the IMU and the associated segment in the OpenSim model, obtained by comparing the initial orientations of the IMU and the segment. As previously described, the segments orientation was derived from the marker-based identification of the model's default pose, which was established while the participant held the adjusted T-pose at the beginning of each trial. \newline

\begin{figure}[htb]
\centering
\includegraphics[width=15cm]{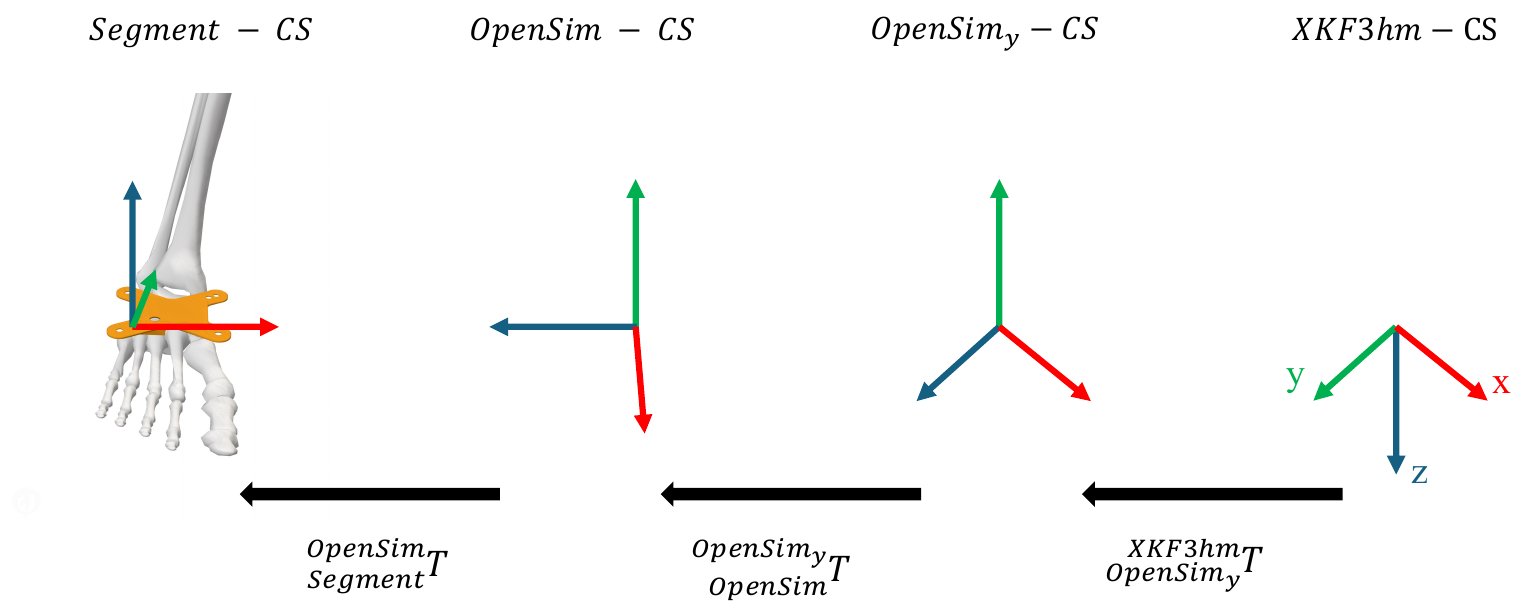}
\caption{Schematic illustration of the transformation chain used to align IMU orientations with their corresponding OpenSim body segments. 
        The transformation $^{\mathrm{XKF3hm}}_{\mathrm{Segment}}T$ is decomposed into three components: 
        a fixed rotation $^{\mathrm{XKF3hm}}_{\mathrm{OpenSim_y}}T$ aligning the vertical axes of the XKF3hm and OpenSim CS; 
        a heading correction $^{\mathrm{OpenSim_y}}_{\mathrm{OpenSim}}T$ accounting for yaw differences between initial IMU and marker-based orientations; 
        and the segment-specific transformation $^{\mathrm{OpenSim}}_{\mathrm{Segment}}T$, derived from the initial pose calibration. 
        All transformations are represented as quaternions.}
\label{fig:transf_chain}
\end{figure}

Both the individual components of the proposed transformation chain and the IMU orientations to which this chain is applied are represented as quaternions (see Figure \ref{fig:transf_chain} for a schematic overview of the transformation process). 

Once the full transformation chain has been applied, the resulting segment-aligned orientations can be used as input to compute joint angles via the \texttt{IMUInverseKinematicsTool}. 
The resulting joint angle trajectories reflect the recorded motion in anatomically interpretable form and provide a valuable basis for subsequent analysis and visualization. \newline
All joint angle trajectories, error estimates, and configuration files required to rerun the Inverse Kinematics procedure, are included in the dataset. 
This allows users to either directly inspect the results or modify and reproduce the processing pipeline according to their own requirements.

\section*{Data Records}

\begin{figure}[htb]
\centering
\includegraphics[width=\linewidth]{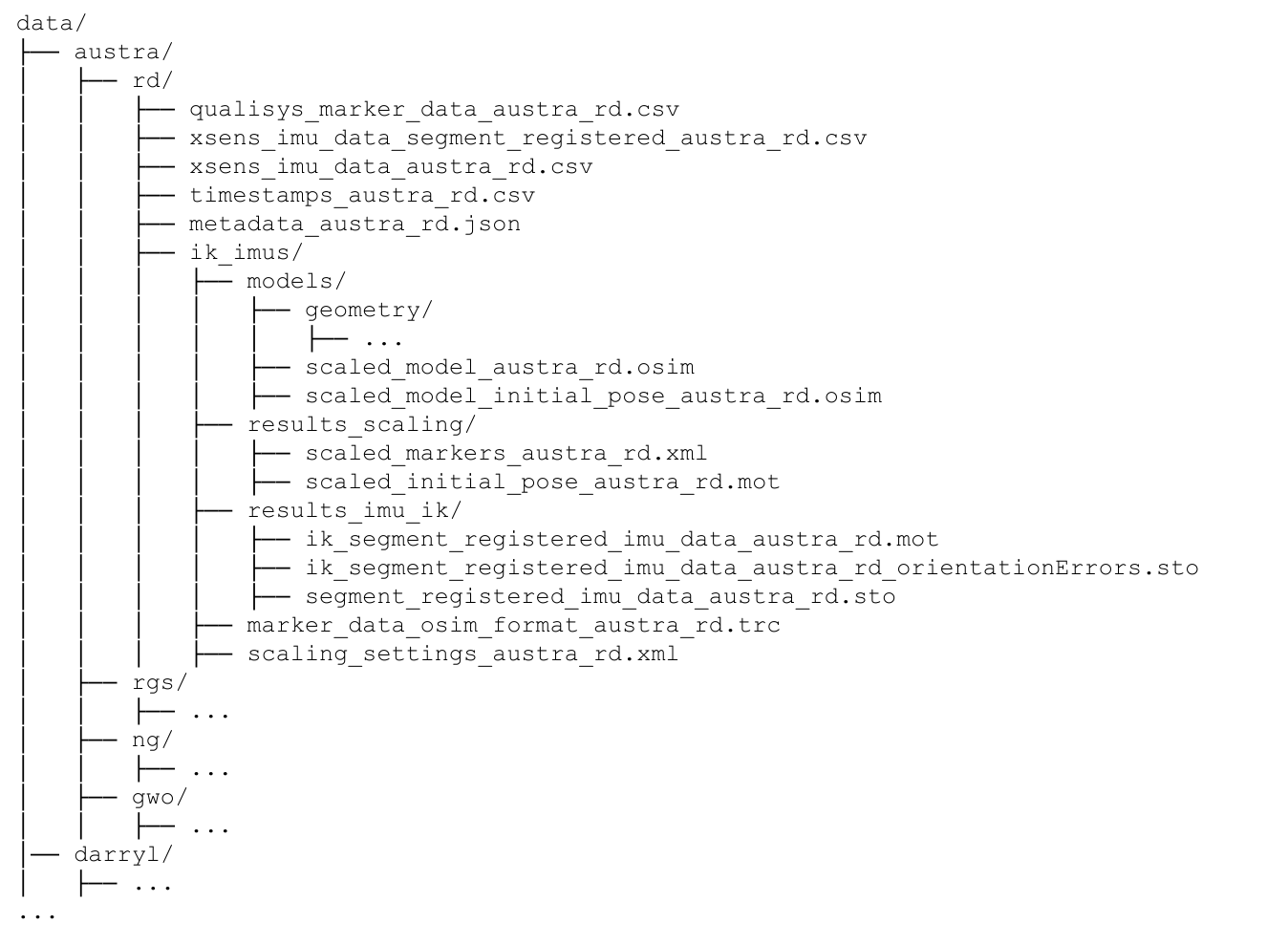}
\caption{Structure of the dataset directory.
Example folder layout for one participant and exercise, including raw data, model files, and results from scaling and Inverse Kinematics.}
\label{fig:folder_tree}
\end{figure}

\modified{The dataset \cite{munzDatasetGAITEXHuman2025} presented in this study is openly available via Zenodo.} 
Its directory structure is depicted in Figure \ref{fig:folder_tree}. 
For each recorded subject, a pseudonym (e.g., austra, darryl) was assigned and used to name a dedicated folder containing the respective data. Within each subject-specific folder, four subfolders were created, each corresponding to one of the exercises performed during the measurement campaign. These subfolders are named using abbreviations of the respective exercise labels (rd, rgs, ng and gwo). The corresponding exercises are explained in detail in Table \ref{tab:exercise_protocol}.

Each exercise-specific folder contains the complete set of data acquired during that particular measurement. 
The 3D marker trajectories recorded with the Qualisys motion capture system are provided in a CSV file following the naming convention \texttt{qualisys\textunderscore marker\allowbreak \textunderscore data\textunderscore <subject>\allowbreak \textunderscore <exercise\textunderscore abbreviation>.csv} (e.g. \texttt{qualisys\allowbreak \textunderscore marker\allowbreak \textunderscore data\textunderscore austra\allowbreak \textunderscore rd.csv}). This file contains the spatial coordinates of all tracked markers over time. The anatomical assignment of marker labels is illustrated in Figure \ref{fig:marker_imu_placement}. \newline
IMU data acquired with Xsens sensors is available in both raw and processed forms. 
The file named \texttt{xsens\allowbreak \textunderscore imu\allowbreak \textunderscore data\allowbreak \textunderscore <subject>\allowbreak \textunderscore <exercise\allowbreak \textunderscore abbreviation>.csv} contains unprocessed orientation data represented as quaternions, as computed by the Xsens system. 
Column names encode both the anatomical sensor location and the respective quaternion component (q, x, y, z), with sensor placements corresponding to the schematic shown in Figure \ref{fig:marker_imu_placement}. 
The processed version \texttt{xsens\allowbreak \allowbreak \allowbreak \textunderscore imu\allowbreak \textunderscore data\allowbreak \textunderscore segment\allowbreak \textunderscore registered\textunderscore <subject>\allowbreak \textunderscore <exercise\allowbreak \textunderscore abbreviation>.csv} contains the same data, with the same column names, after applying the transformation chain described in Equation \ref{eq:transf}, aligning all sensor orientations with the local CS of the associated body segments of the OpenSim model. \newline
Each folder also includes a \texttt{timestamps\allowbreak \textunderscore <subject>\allowbreak \textunderscore <exercise\allowbreak \textunderscore abbreviation>.csv} file, which defines the start and end times of each repetition or, for gait tasks, the timing of different walking speed conditions.
The column names use the aliases defined in Table \ref{tab:exercise_protocol} to encode the performed exercise and variant.\newline
Each of the above files contain a complete measurement trial, including all repetitions performed by a given subject for the respective task (e.g., all executions of the Resisted Dorsiflexion exercise).\newline
Additional metadata is stored in \texttt{metadata\allowbreak \textunderscore <subject>\allowbreak \textunderscore <exercise\allowbreak \textunderscore abbreviation>.json}, which contains subject-specific parameters relevant for automated model personalization and reproducibility. 
Specifically, the file allows to define manual mappings between sensor names and body segments (to accommodate variations in sensor labeling) and record exclusions of specific segments from scaling, for instance, if markers were missing or unreliable during the static pose. 
These settings are critical when using the tools provided in the associated GitHub repository, which expects this metadata format for automated scaling and Inverse Kinematics processing.

Each exercise-specific folder also contains a subdirectory named \texttt{ik\allowbreak \textunderscore imus}, which holds all files produced during subject-specific model scaling and the execution of IMU-based Inverse Kinematics.
The \texttt{models} subfolder contains subject-specific musculoskeletal models in OpenSim format. 
The file \texttt{scaled\allowbreak \textunderscore model\allowbreak \textunderscore <subject>\allowbreak \textunderscore <exercise\allowbreak \textunderscore abbreviation>\allowbreak.osim} represents the scaled version of the generic model, whose segment dimensions have been adjusted to match the subject based on optical marker data.
A corresponding file with the naming convention \texttt{scaled\allowbreak \textunderscore model\allowbreak \textunderscore initial\allowbreak \textunderscore pose\allowbreak \textunderscore <subject>\allowbreak \textunderscore <exercise\allowbreak \textunderscore abbreviation>.osim} includes the subject’s initial T-pose derived from the marker data.\newline
The subfolder \texttt{results\allowbreak \textunderscore imu\allowbreak \textunderscore ik} subdirectory contains the output of the Inverse Kinematics procedure using segment-registered IMU data. 
The file \texttt{ik\allowbreak \textunderscore segment\allowbreak \textunderscore registered\allowbreak \textunderscore imu\allowbreak \textunderscore data\allowbreak \textunderscore <subject>\allowbreak \textunderscore <exercise\allowbreak \textunderscore abbreviation>\allowbreak.mot} provides the estimated joint angles over time and can be used for visualization within OpenSim, together with the corresponding model. 
The file \texttt{ik\allowbreak \textunderscore segment\allowbreak \textunderscore registered\allowbreak \textunderscore imu\allowbreak \textunderscore data\allowbreak \textunderscore <subject>\allowbreak \textunderscore <exercise\textunderscore abbreviation>\allowbreak \textunderscore orientation\allowbreak \textunderscore Errors.sto} reports the orientation discrepancies between the experimental IMU data and the corresponding models segment, that remained after optimization. 
The input to the Inverse Kinematics process is provided in \texttt{segment\allowbreak \textunderscore registered\allowbreak \textunderscore imu\textunderscore data\allowbreak \textunderscore <subject>\allowbreak \textunderscore <exercise\allowbreak \textunderscore abbreviation>.sto}, which is a reformatted version of the CSV file \texttt{xsens\allowbreak \textunderscore imu\allowbreak \textunderscore data\allowbreak \textunderscore segment\allowbreak \textunderscore registered\allowbreak \textunderscore <subject>\allowbreak \textunderscore <exercise\textunderscore abbreviation>.csv}, structured for compatibility with OpenSim’s \texttt{IMUInverseKinematicsTool}. \newline
The results of the model scaling process are stored in the  \texttt{results\allowbreak \textunderscore scaling} subfolder. The file \texttt{scaled\allowbreak \textunderscore initial\allowbreak \textunderscore pose\allowbreak \textunderscore <subject>\allowbreak \textunderscore <exercise\allowbreak \textunderscore abbreviation>.mot} defines the model's initial pose so that the scaled model can be correctly positioned for subsequent analyses. 
The file \texttt{scaled\allowbreak \textunderscore markers\textunderscore <subject>\allowbreak \textunderscore <exercise\allowbreak \textunderscore abbreviation>.xml} contains the results of the marker-based scaling procedure, including marker placement and scaling factors.

To support reproducibility and reuse, additional configuration files are included in the \texttt{ik\textunderscore imus} subfolder. 
The file \texttt{marker\allowbreak \textunderscore data\textunderscore osim\allowbreak \textunderscore format\textunderscore <subject>\allowbreak \textunderscore <exercise\textunderscore abbreviation>.trc} provides a TRC-formatted version of the Qualisys marker trajectories for direct use in OpenSim workflows. 
The file \texttt{scaling\allowbreak \textunderscore settings\textunderscore <subject>\allowbreak \textunderscore <exercise\allowbreak \textunderscore abbreviation>.xml} contains all parameter settings used during scaling, enabling users to replicate the model generation step programmatically using OpenSim’s command-line or GUI-based tools.

\section*{Technical Validation}

To verify the integrity of the dataset, we conducted \modified{a technical validation} addressing three central objectives: (i) to assess whether the applied transformation correctly maps each IMU’s orientation from its local CS into the OpenSim global reference frame; \modified{(ii) to offer exemplary support for a successful temporal synchronization between the Xsens and Qualisys systems} and (iii) to quantify how the orientation estimates provided by the Xsens system deviate over time from the marker-based orientation of each IMU, as determined via the Qualisys system. This analysis thus provides a robust validation of both spatial and temporal alignment and allows users to track potential drift or systematic discrepancies in orientation estimation throughout each recording.

Each IMU was equipped with a rigidly mounted marker plate carrying four Qualisys markers, allowing us to derive the orientation of the IMU in the global Qualisys CS. 
Since the Qualisys and OpenSim CS are congruent by design, this marker-based orientation, denoted as $^{Marker}_{OpenSim}T$, is expressed in the OpenSim frame.
Additionally, the estimated orientation of each IMU was transformed to align with the global Qualisys CS via the following transformation chain: 

\begin{equation}
^{IMU}_{OpenSim}T = ^{OpenSim_y}_{OpenSim}T\cdot^{XKF3hm}_{OpenSim_y}T
\label{eq:transf2}
\end{equation}

This expression (Equation \ref{eq:transf2}) is based on the general transformation structure introduced in Equation \ref{eq:transf} and maps the IMU orientation from the internal XKF3hm into the global OpenSim/Qualisys frame. After applying this transformation chain to the recorded IMU orientation, both the Xsens-based and marker-based orientation estimates reside in the same reference frame and are directly comparable. All used orientations and transformations were represented as quaternions.

The relative orientation between the two was then computed as:

\begin{equation}
^{IMU}_{Marker}T = ^{OpenSim}_{Marker}T\cdot^{IMU}_{OpenSim}T
\label{eq:transf3}
\end{equation}

This transformation captures the deviation between the two orientation estimates. We converted it into axis-angle representation and extracted the angular magnitude to quantify the deviation.

\begin{figure}[htb]
\centering
\includegraphics[width=\linewidth]{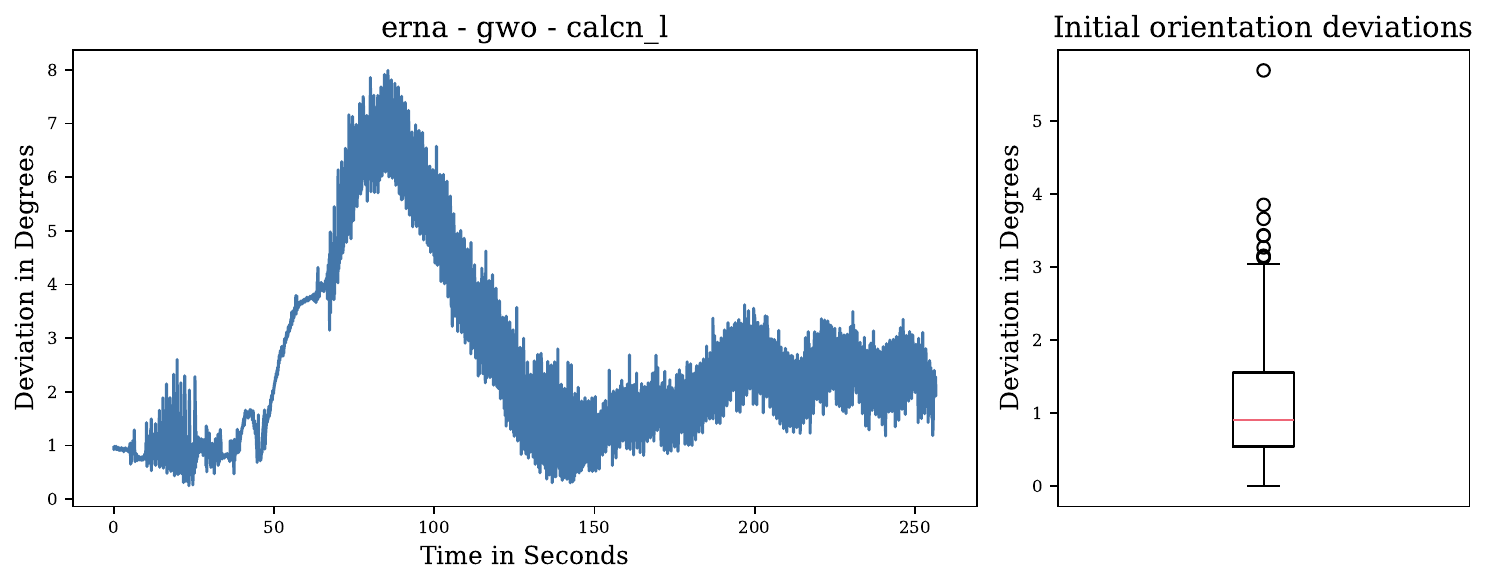}
\caption{Left: Temporal evolution of the orientation deviation between Xsens- and Qualisys-based estimates for a representative trial. Right: Overview of the initial orientation offsets across all recordings. The deviation is quantified as the magnitude of the axis-angle representation of the relative orientation between the two systems, computed in the global OpenSim frame.}
\label{fig:orientation_dev}
\end{figure}

The left panel of Figure \ref{fig:orientation_dev} illustrates the angular deviation for a representative IMU during a single trial. Initially, the deviation remains low, indicating that the transformation chain in Equation \ref{eq:transf2} and consequently the corresponding part in Equation \ref{eq:transf}, accurately maps the IMU data into the OpenSim frame. The right panel of Figure \ref{fig:orientation_dev} summarizes the initial orientation deviation across all IMUs and trials. The vast majority of deviations lie within 0 to 3.15 degrees, indicating a high level of agreement between the two orientation estimates. The remaining discrepancies are likely attributable to subtle inconsistencies in the definition of the vertical axis. Specifically, while the Xsens system determines the global “down” direction based on the measured gravitational vector and internal sensor calibration, the marker-based definition of the vertical axis in the Qualisys system may differ slightly. These differences could arise either from small deviations in the alignment of the accelerometer within the IMU housing or from an imperfectly parallel calibration reference used to define the Qualisys CS.

Furthermore, the left panel of Figure \ref{fig:orientation_dev} offers empirical support for successful temporal synchronization. If time synchronization were flawed, one would expect sharp, transient spikes in the angular deviation. However, such abrupt deviations are not observed, suggesting that temporal alignment between Xsens and Qualisys data is maintained throughout the recordings.

Moreover, the continuous evolution of the deviation angle across time offers insights into the stability of Xsens orientation estimates. In many trials, a gradual increase in deviation was observed, which we attribute to heading drift in the Xsens estimation.

Although not all observed discrepancies could be visualized within the scope of this paper, we provide a fully documented analysis script in the accompanying GitHub repository. This allows users to replicate our calculations, visualize orientation discrepancies, and adapt the tools for extended analysis of spatial or temporal inconsistencies.

\section*{Usage Notes}
As is common in experimental studies, occasional minor deviations from the intended measurement protocol occurred during the course of the data collection campaign. 
To ensure full transparency, we provide a dedicated file (UsageNotes.md) in which all protocol deviations are documented on a per-participant basis. 
These deviations do not affect the integrity of the recorded sensor data, but may include issues such as missing Qualisys markers or minor procedural variations, for example, cases where the initial T-pose was assumed only after an explicit prompt rather than immediately at the start of the recording.

\modified{In addition, one specific technical issue should be carefully considered when working with the IMU orientation data. 
Post-hoc analysis revealed a deviation in the magnetometer calibration of two IMUs (\texttt{XSens\textunderscore Lower\textunderscore Leg\textunderscore Right} and \texttt{XSens\textunderscore Hand\textunderscore Right}). 
In contrast to the other units, this IMU used a different reference frame for heading estimation, resulting in a systematic yaw offset. This discrepancy is accounted for in the transformation step $^{OpenSim_y}_{OpenSim}T$ (as defined in equation \ref{eq:transf}) applied during the processing pipeline.
Accordingly, this issue is already compensated for in the processed orientation trajectories
(\texttt{xsens\allowbreak \textunderscore imu\allowbreak \textunderscore data\allowbreak \textunderscore segment\allowbreak \textunderscore registered\allowbreak \textunderscore <subject>\allowbreak \textunderscore <exercise\textunderscore abbreviation>.csv}),
but not in the raw data
(\texttt{xsens\allowbreak \textunderscore imu\allowbreak \textunderscore data\allowbreak \textunderscore <subject>\allowbreak \textunderscore <exercise\textunderscore abbreviation>.csv}).
The required compensation transformation $^{OpenSim_y}_{OpenSim}T$ is computed individually for each recording.
The exact value of this deviation can be determined for every measurement using the provided raw data and the processing scripts available in the associated repository.
}

\section*{Code Availability}
All post-processing steps described in this manuscript—including model scaling, orientation transformation, and inverse kinematics—can be reproduced using the accompanying code repository. 
The repository is publicly available on Github via the following URL (\url{https://github.com/ai-for-sensor-data-analytics-ulm/aisd_ortho_ki_dataset}) and provides scripts, configuration files, and documentation required to apply the full processing pipeline to the provided raw data.


\section*{Author contributions statement}

Conceptualization, A.S., H.O., J.W., K.S.-S., F.C. and M.M.; Ethics Approval, K.S.-S., F.C. and M.M.; Data Collection, A.S., H.O. and J.W.;  Data Curation, A.S. and H.O.;  Postprocessing, A.S. and H.O.; Resources, F.C. and M.M.; Software, A.S. and H.O.;  Validation, A.S., H.O. and M.M.; Visualization, A.S., H.O. and J.W.; Supervision, M.M.; Project administration, M.M.; Funding acquisition, F.C. and M.M.; Writing-original draft, A.S. and H.O.; Writing-review and editing, A.S., H.O. and M.M. All authors have read and agreed to the published version of the manuscript.

\section*{Competing Interests}
The authors declare no competing interests.

\section*{Acknowledgements}
The OrthoKI project is funded by the Carl-Zeiss-Stiftung.

\end{document}